\documentclass[sigconf]{acmart}
\AtBeginDocument{%
  }

\usepackage{booktabs}
\usepackage{graphicx}
\usepackage{amsmath}
\usepackage{multirow}
\usepackage{xcolor}
\usepackage{tikz}
\usepackage{enumitem}
\graphicspath{{figures/}}

\newcommand{\best}[1]{\textbf{#1}}
\newcommand{\sys}{\textsc{ArtSociety}}
\newcommand{\dtoc}{DESCRIBE\discretionary{}{}{}$\,\rightarrow\,$\discretionary{}{}{}CLASSIFY}
\newcommand{\caseimg}[1]{%
  \begin{minipage}[c][2.5cm][c]{0.96\linewidth}%
    \centering\includegraphics[width=0.96\linewidth,height=2.4cm]{#1}%
  \end{minipage}%
}
\newcommand{\casetxt}[1]{{\scriptsize\linespread{0.88}\selectfont
  \setlength{\emergencystretch}{1.5em}\hyphenpenalty=50 #1\par}}

\copyrightyear{2026}
\acmYear{2026}
\setcopyright{cc}
\setcctype{by}
\acmConference[MM '26]{Proceedings of the 34th ACM International Conference on Multimedia}{November 10--14, 2026}{Rio de Janeiro, Brazil}
\acmBooktitle{Proceedings of the 34th ACM International Conference on Multimedia (MM '26), November 10--14, 2026, Rio de Janeiro, Brazil}
\acmDOI{10.1145/3767308.3837723}
\acmISBN{979-8-4007-2213-4/2026/11}

\begin{document}

\title{ArtSociety: Multi-Agent Multimodal Collaboration for Art Emotion Understanding}

\author{Jian Li}
\affiliation{%
  \institution{Nanjing University}
  \city{Suzhou}
  \country{China}
}
\affiliation{%
  \institution{Tencent YoutuLab}
  \city{Shanghai}
  \country{China}
}
\email{swordlidev@gmail.com}

\author{Fanfan Ji}
\affiliation{%
  \institution{Nanjing University}
  \city{Suzhou}
  \country{China}
}
\email{jiff1995@nju.edu.cn}

\author{Jinxiang Lai}
\affiliation{%
  \department{CSE}
  \institution{The Hong Kong University of Science and Technology}
  \city{Hong Kong}
  \country{China}
}
\email{layjins1994@gmail.com}

\author{Ying Tai}
\affiliation{%
  \institution{Nanjing University}
  \city{Suzhou}
  \country{China}
}
\email{yingtai@nju.edu.cn}

\author{Jian Yang}
\affiliation{%
  \institution{Nanjing University}
  \city{Suzhou}
  \country{China}
}
\email{csjyang@nju.edu.cn}

\author{Xiao-Tong Yuan}
\authornote{Corresponding author.}
\affiliation{%
  \institution{Nanjing University}
  \city{Suzhou}
  \country{China}
}
\email{xtyuan@nju.edu.cn}

\author{Chengjie Wang}
\affiliation{%
  \institution{Tencent YoutuLab}
  \city{Shanghai}
  \country{China}
}
\email{jasoncjwang@tencent.com}

\author{Yabiao Wang}
\affiliation{%
  \institution{Tencent YoutuLab}
  \city{Shanghai}
  \country{China}
}
\email{caseywang@tencent.com}

\renewcommand{\shortauthors}{Jian Li et al.}
\begin{abstract}
The AffectiveArt Multidimensional Art Emotion Understanding asks to jointly predict an artwork's
fine-grained emotion (12 classes, $1549{:}1$ head-to-tail ratio), binary valence/arousal,
and five attribute-grounded descriptions---sub-tasks that exhibit strong empirical trade-offs,
so the single-model solutions we tried do not jointly optimize all of them well.
We present \sys{}, a multi-agent framework that assembles \textbf{heterogeneous multimodal
experts}---a DINOv2-Giant vision agent ($A_1$), a scene-grounded CoT fine-tuned MLLM ($A_2$),
and three closed-source reasoning agents ($A_3$--$A_5$)---and coordinates them with two
training-free controllers: (i)~a \emph{rare-class-aware voting arbiter} that lowers the
agreement threshold for tail emotions, exploiting decorrelated error patterns across agent
families; and (ii)~a \emph{description-first reasoning agent} whose \dtoc{} chain of thought
forces visual evidence before label commitment, yielding near-perfect grounded descriptions.
A task-routing policy directs the hard emotion task to the full five-agent ensemble while
assigning the near-saturated V/A and generative description tasks to the single strongest
reasoning agent. On the official test set ($1{,}000$ artworks), \sys{} achieves an
\textbf{Overall Score of $0.8870$} (Classification $0.7789$, Description $0.9952$).
An eleven-variant ablation study reveals that, once method and scale saturate at
${\sim}0.76$, the decisive gains come from \emph{agent collaboration} and \emph{data-side
supervision}---a $30$B MoE model trained on older data does not outperform an $8$B model trained on better data.
Code is available at \url{https://github.com/swordlidev/ArtSociety}.
\end{abstract}



\maketitle

\section{Introduction}

Understanding the emotional content of visual art is a long-standing challenge at the
intersection of affective computing, computer vision, and art
history~\cite{machajdik2010affective,achlioptas2021artemis}. Unlike natural images whose
affective cues are often tied to facial expressions or scene semantics, artworks convey
emotion through abstract formal elements---brushstroke energy, color harmony, compositional
tension, and lighting mood---requiring both perceptual sensitivity and art-historical
reasoning.

The \textbf{AffectiveArt 2026 Grand Challenge, Track~2} formalizes this problem by asking
systems to jointly predict an artwork's dominant emotion (12 categories), binary valence and
arousal, and five attribute-grounded descriptions tied to formal qualities (brushstroke,
composition, color, line, light)~\cite{yang2023emoset,zhang2025emoart}. Submissions are
scored as the equal-weight average of a classification score and a description score assessed
by a multimodal LLM protocol. This multi-output formulation demands that a system
simultaneously excel at discrete classification, dimensional prediction, and open-ended
natural-language generation---a combination that single-model architectures struggle to
optimize end-to-end.

Analysis of EmoArt-130k~\cite{zhang2025emoart} reveals three structurally different
difficulties. \textbf{(1)~Long-tailed emotion distribution:} the label space is severely
imbalanced ($1549{:}1$ head-to-tail ratio, Gini $=0.741$), making accuracy and macro-F1
trade off sharply and rendering standard loss re-weighting inadequate. \textbf{(2)~Near-%
saturated binary axes:} valence and arousal already exceed $90\%$ accuracy with naive
baselines, leaving little headroom for ensemble gains. \textbf{(3)~Grounded description as
reasoning:} the description sub-task rewards art-historical vocabulary and evidence-anchored
explanation rather than label accuracy, favoring large-language-model generation over
discriminative classifiers. A single model optimized for one regime is fundamentally
mis-matched to the others.

These observations motivate our design: an \textbf{agentic society of specialized experts}.
Rather than training one monolithic model, we assemble heterogeneous
agents---DINOv2~\cite{oquab2024dinov2} for visual feature extraction,
Qwen3-VL~\cite{qwen3vl2025} fine-tuned with scene-grounded chain-of-thought,
GPT-5.4~\cite{openai2025gpt5} for zero-shot reasoning with prior knowledge, and
Gemini~3.1~\cite{team2023gemini} for stylistic diversity---and coordinate them with two
training-free controllers.
A \emph{rare-class-aware voting arbiter} resolves the 12-way emotion task by aggregating
votes across all agents with a lowered confidence threshold for rare classes, exploiting
complementary error patterns among diverse model families.
A \emph{description-first reasoning agent} produces valence, arousal, and grounded
descriptions via a \dtoc{} chain of thought that forces the model to observe visual evidence
before committing to emotional labels.
A task-routing policy assigns the hard emotion task to the full ensemble while directing the
saturated/generative tasks to the single best reasoning agent, minimizing latency without
sacrificing quality.

An eleven-variant ablation study (\S\ref{sec:exp}) reveals a key insight: method- and
scale-side changes plateau at ${\sim}0.76$ Combined, and a $30$B MoE trained on older data does \emph{not} beat an
$8$B model trained on better data. The decisive gains come from (a)~genuine CoT supervision
that teaches structured visual reasoning and (b)~heterogeneous agent collaboration that
captures diverse failure modes---reframing the contest as a problem of \emph{collaboration
and data quality}, not raw model capacity.
We present \sys{}, a multi-agent framework for art emotion understanding that unifies heterogeneous expert agents, task-aware routing, and structured visual reasoning. Our main contributions are:
\begin{itemize}[leftmargin=1.4em,itemsep=1pt,topsep=2pt]
  \item We formulate art emotion understanding as an \textbf{agentic society of multimodal
  experts} with task-aware routing, matching each sub-task to a tailored agent configuration.
  \item We introduce a \textbf{rare-class-aware voting arbiter} that exploits complementary
  errors across five heterogeneous agents, improving tail-class recognition accuracy.
  \item A \textbf{description-first reasoning agent}
  yields grounded descriptions ($0.9952$) with \dtoc{} ordering.
  \item An \textbf{eleven-variant ablation study} shows that data-side supervision and
  agent collaboration drive gains past method/scale saturation. Our system achieves
  \textbf{Overall $0.8870$} (Cls.\ $0.7789$, Desc.\ $0.9952$) on the official test set,
  achieving the \textbf{top public leaderboard score} on the AffectiveArt 2026 Track~2 leaderboard at the time of submission (team N\&T).
\end{itemize}
\section{Related Work}

\subsection{Art Emotion Analysis and MLLMs}
Visual emotion analysis evolved from hand-crafted
features~\cite{machajdik2010affective,borth2013large} to deep-learning
datasets~\cite{you2016building,mollahosseini2017affectnet,mertens2024findingemo}. In the art
domain, ArtEmis~\cite{achlioptas2021artemis,mohamed2022artemisv2} introduced affective
captioning over WikiArt, EmoSet~\cite{yang2023emoset} added rich attributes, and
EmoArt-130k~\cite{zhang2025emoart} first combined fine-grained emotion labels, binary
valence/arousal (Russell's circumplex~\cite{russell1980circumplex}), and structured
descriptions---underlying the AffectiveArt challenge. On the modeling side, multimodal LLMs
(MLLMs) such as EmoLLM~\cite{yang2024emollm}, Emotion-LLaMA~\cite{cheng2024emotionllama},
AffectGPT~\cite{lian2025affectgpt}, and EmoVLM-KD~\cite{cao2025emovlmkd} advance emotion
recognition and reasoning, yet evaluations of GPT-4V/Gemini-class
models~\cite{lu2024mllm} reveal strong captioning but unstable fine-grained
classification---the asymmetry we exploit by routing different sub-tasks to different agents.

\subsection{Attribute-Grounded Reasoning and Multi-Agent Collaboration}
Grounding affective judgments in visual attributes is central to recent
work: AMH-Net~\cite{amhnet2026} fuses MLLM-generated descriptions with visual features, and
AGSR/FAB-G~\cite{zhang2026agsr} uses a supervised multi-agent pipeline to select emotionally
operative attributes. Our system shares the attribute-grounded philosophy but differs in using
\emph{heterogeneous models} coordinated by a training-free voting arbiter rather than
role-prompts of one model. Chain-of-thought prompting~\cite{wei2022cot} and
self-consistency~\cite{wang2023selfconsistency} inform our \dtoc{} reasoning order.
More broadly, multi-agent debate~\cite{du2023debate} and
mixture-of-agents~\cite{wang2024mixtureofagents} improve LLM factuality; we adapt these ideas
to a multimodal, long-tailed setting with an explicit rare-class arbiter---addressing extreme
imbalance ($1549{:}1$) where focal loss~\cite{lin2017focal} and class-balanced
loss~\cite{cui2019classbalanced} prove insufficient. Fine-tuning uses
LoRA~\cite{hu2022lora} via LlamaFactory~\cite{zheng2024llamafactory}.
We are the first to unify heterogeneous MLLMs into a collaborative society with task-aware routing and a rare-class-aware arbiter for art emotion understanding.

\section{Problem and Data Analysis}
\label{sec:problem}

\begin{figure}[t]
 \vspace{-6pt}
  \centering
  \includegraphics[width=\columnwidth]{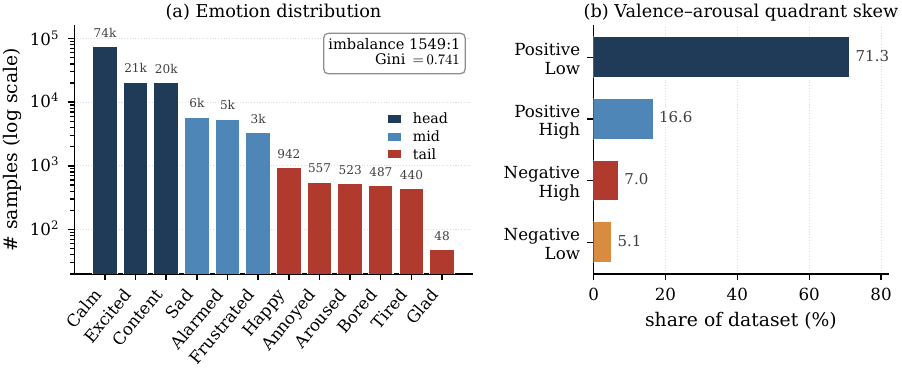}
  \caption{EmoArt-130k data characteristics. \textbf{(a)} Emotion distribution (log scale): $1549{:}1$ imbalance makes tail classes decisive. \textbf{(b)} V/A quadrant skew: Positive-Low covers $71.3\%$; binary V/A is near-saturated.}
  \label{fig:dist}
   \vspace{-6pt}
\end{figure}

\subsection{Task Formalization}
Given an artwork image $x$, a system must output a tuple
$y=(e, v, a, c, \{d_k\}_{k=1}^{5})$, where $e\in\mathcal{E}$ is one of $12$ emotion categories
$\mathcal{E}=\{$\textsc{aroused, excited, happy, alarmed, annoyed, frustrated, sad, bored,
tired, content, calm, glad}$\}$; $v\in\{$Positive, Negative$\}$ is the valence;
$a\in\{$High, Low$\}$ is the arousal; $c$ is a holistic emotional caption; and
$\{d_k\}$ are five attribute analyses for \emph{brushstroke}, \emph{composition},
\emph{color}, \emph{line}, and \emph{light}. The official Track-2 score is
\begin{equation}
S = \tfrac{1}{2}\,S_{\text{cls}} + \tfrac{1}{2}\,S_{\text{desc}},
\end{equation}
where the classification score averages the three label sub-tasks, each scored as the equal
weight of macro-F1 and accuracy,
\begin{equation}
S_{\text{cls}} = \tfrac{1}{3}\!\!\sum_{t\in\{e,v,a\}}\!\!\Big(\tfrac{1}{2}\text{F1}_t +
\tfrac{1}{2}\text{Acc}_t\Big),
\end{equation}
and $S_{\text{desc}}$ aggregates artwork consistency, attribute-analysis quality, and
overall-caption quality via a standardized multimodal LLM-assisted protocol. The equal weight
on macro-F1 makes rare-class performance decisive: ignoring the tail caps $S_{\text{cls}}$
regardless of accuracy.

\subsection{Dataset Characteristics and Three Difficulty Regimes}
We analyze the official EmoArt-130k training corpus ($132{,}895$ artworks, $56$ painting
styles). Three properties shape our design.

\noindent\textbf{(1) Extreme emotion long-tail.}
Figure~\ref{fig:dist}(a) shows the emotion distribution on a log scale. \textsc{Calm}
($74{,}350$) dominates at $55.95\%$---$3.6\times$ the runner-up \textsc{Excited}---while the
six rarest classes together account for only $2.26\%$ and \textsc{Glad} has merely $48$
examples. The imbalance ratio is $1549{:}1$ and the Gini coefficient is $0.741$. Consequently a
degenerate majority predictor scores $\sim\!56\%$ accuracy but near-zero macro-F1, so accuracy
and macro-F1 trade off sharply.

\noindent\textbf{(2) Near-saturated valence/arousal.}
Valence is $87.9\%$ positive and arousal is $76.4\%$ low; the four V-A quadrants are highly
skewed (Figure~\ref{fig:dist}(b)), with \emph{Positive-Low} alone covering $71.3\%$. Yet because
these axes are binary and visually salient, strong models reach $>\!90\%$ accuracy with little
effort---headroom here is small.

\noindent\textbf{(3) A reasoning task, plus a data ceiling.}
Attribute-grounded description rewards faithful observation and art vocabulary rather than
label accuracy. Moreover, manual inspection reveals that for the rarest classes (e.g.,
\textsc{Glad}, \textsc{Tired}, \textsc{Annoyed}) the free-text emotional-impact annotations are
frequently \emph{semantically inconsistent} with the assigned label, an artifact of the
automatic annotation stage. This imposes a \emph{data ceiling}: no method we tried recovers a
stable visual$\rightarrow$label mapping for these classes.

In our experiments, these regimes are difficult to optimize jointly within a single model:
pushing rare-class recall hurts majority accuracy, optimizing classification does not improve
description, and scaling capacity hits the data ceiling. This motivates the agentic decomposition of
\S\ref{sec:method}.

\section{Method: A Multi-Agent Society of Experts}
\label{sec:method}

\subsection{Overview}
\begin{figure}[t]
\vspace{-6pt}
\centering
\includegraphics[width=0.95\columnwidth]{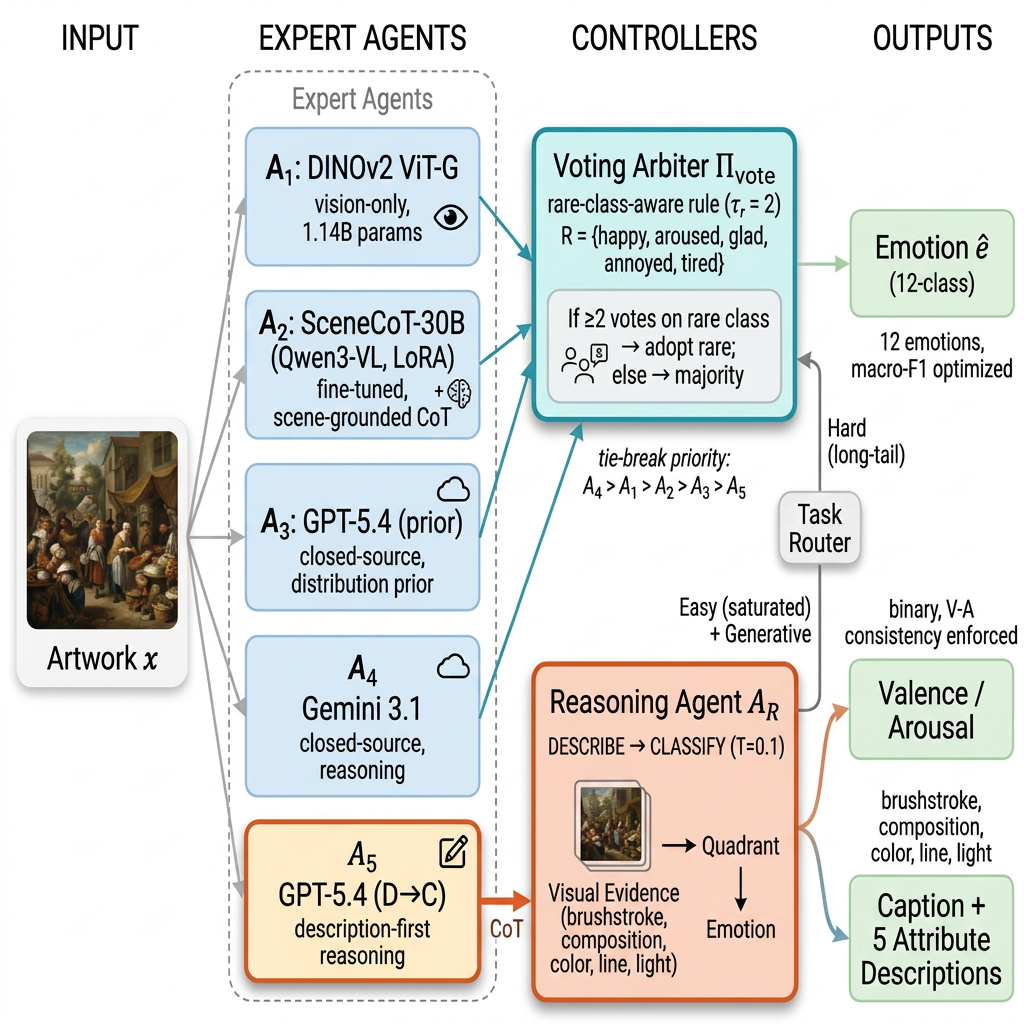}
\caption{The \sys{} agentic framework. Five expert agents emit emotion votes resolved by arbiter $\Pi_{\text{vote}}$; the reasoning agent $A_5$ produces valence, arousal, and grounded descriptions. Artwork images are from EmoArt-130k~\cite{zhang2025emoart} }
 \vspace{-6pt}
\label{fig:arch}
 \vspace{-6pt}
\end{figure}

Our system, \sys{}, treats the multi-output challenge as a \emph{multi-agent society of
specialized experts} coordinated by two training-free controllers and a task-routing policy
(Figure~\ref{fig:arch}). The key insight is that the three difficulty regimes identified in
\S\ref{sec:problem} call for fundamentally different agent capabilities: long-tailed emotion
classification benefits from \emph{diverse, decorrelated voters}; near-saturated V/A prediction
needs only one strong model; and grounded description demands \emph{reasoning depth}, not
ensemble breadth. Formally, a set of \emph{expert agents}
$\mathcal{A}=\{A_1,\dots,A_5\}$ each map an artwork $x$ to a partial prediction; a
\emph{voting arbiter} $\Pi_{\text{vote}}$ aggregates their emotion votes into $\hat e$; a
\emph{reasoning agent} $A_R{\equiv}A_5$ produces valence $\hat v$, arousal $\hat a$, and the five grounded
descriptions $\{\hat d_k\}$ and caption $\hat c$ via chain-of-thought; and a router assigns
each sub-task to the configuration that matches its difficulty regime.
The guiding principle is \emph{diversity over capacity}: agents are
chosen to have complementary error patterns rather than to be individually optimal.

\subsection{Expert Agents}
We span four model paradigms so that errors decorrelate.

\noindent\textbf{$A_1$ -- Self-supervised vision agent.} A DINOv2
ViT-Giant backbone~\cite{oquab2024dinov2} ($1.14$B params, $518{\times}518$
input) with lightweight classification heads, fine-tuned on EmoArt with
$\sqrt{\cdot}$ inverse-frequency class weights and balanced re-sampling
(Calm under-sampled to ${\sim}30\%$). It carries \emph{no language prior},
so its mistakes are visual rather than semantic, complementing the MLLM agents.

\noindent\textbf{$A_2$ -- Fine-tuned open MLLM agent.} We LoRA-fine-tune~\cite{hu2022lora}
Qwen3-VL~\cite{qwen3vl2025} with LlamaFactory~\cite{zheng2024llamafactory} (rank $128$,
$\alpha{=}256$, vision tower and multi-modal projector unfrozen), producing a structured CoT
JSON with $100\%$ parse reliability. The training targets are constructed from EmoArt-130k's
own annotations: each sample's \emph{scene description} and \emph{emotional-impact reasoning}
(from the dataset's human-curated fields) are assembled into a four-stage CoT target
(scene $\to$ attributes $\to$ V-A reasoning $\to$ emotion), with style$\times$emotion joint
stratification to eliminate validation bias.
The final voter uses SceneCoT-30B, a $30$B MoE variant
trained on this scene-grounded CoT data---the best single-agent configuration identified in
\S\ref{sec:exp}.

\noindent\textbf{$A_3,A_4$ -- Closed-source reasoning agents.} GPT-5.4~\cite{openai2025gpt5}
and Gemini~3.1~\cite{team2023gemini}, prompted with a reasoning chain and a distribution prior.
Coming from different families, they contribute predictions uncorrelated with the open agents.

\noindent\textbf{$A_5$ -- Description-first reasoning agent.} A specialized GPT-5.4
configuration that doubles as a fifth voter and as the
reasoning agent for V/A and descriptions (\S\ref{sec:reason}). Its prompt encodes:
(i)~a structured emotion taxonomy organized by Russell's
circumplex quadrants~\cite{russell1980circumplex} with pairwise differentiation criteria;
(ii)~quadrant-first reasoning (decide valence $\to$ arousal $\to$ quadrant $\to$ best
in-quadrant emotion); and
(iii)~describe-before-classify field ordering, forcing visual grounding before label
commitment. A post-hoc consistency check auto-corrects V-A/emotion mismatches. Decoding
uses $T{=}0.1$ with quality-aware retries on malformed output.

\subsection{Rare-Class-Aware Voting Arbiter}
The hardest sub-task, $12$-way emotion, is resolved by $\Pi_{\text{vote}}$ over the five
agents' votes $\{e^{(1)},\dots,e^{(5)}\}$. Let $n(e)=\sum_i \mathbb{1}[e^{(i)}=e]$ be the vote
count for class $e$. Standard majority voting assigns the head class whenever it appears, which
re-introduces the very majority bias we fight. We therefore use an asymmetric, rare-class-aware
rule:
\begin{equation}
\hat e =
\begin{cases}
\arg\max_{e\in\mathcal{R}} n(e), & \text{if } \max_{e\in\mathcal{R}} n(e)\ge \tau_r,\\[2pt]
\arg\max_{e\notin\mathcal{R}} n(e), & \text{otherwise},
\end{cases}
\end{equation}
where $\mathcal{R}=\{$\textsc{happy, aroused, glad, annoyed, tired}$\}$ is the rare set and the
threshold $\tau_r=2$ (vs.\ the implicit $\ge 3$ for common classes). Intuitively, if two
independent agents both ``see'' a rare emotion, that agreement is strong evidence and should
override a larger but lower-precision majority block. Ties are broken by a fixed priority order
calibrated on validation macro-F1: $A_4$(Gemini~3.1) $>$ $A_1$(DINOv2) $>$ $A_2$(Qwen3-VL) $>$
$A_3$(GPT-5.4, prior) $>$ $A_5$(GPT-5.4, D$\to$C). This single mechanism lifts test-set emotion accuracy
to $0.796$, the best among all configurations, while improving macro-F1 over any single agent.

\subsection{Description-First Reasoning Agent}
\label{sec:reason}
Valence, arousal, and all textual outputs are produced by $A_5$ under a strict
\dtoc{} chain of thought~\cite{wei2022cot}:
\begin{enumerate}[leftmargin=1.5em,itemsep=1pt,topsep=2pt]
  \item \textbf{Describe first.} The output JSON schema places the five attribute fields
  (brushstroke, composition, color, line, light) and a scene description \emph{before} any
  label, forcing the agent to verbalize visual evidence before committing to an emotion.
  \item \textbf{Quadrant-first labeling.} Following Russell's circumplex~\cite{russell1980circumplex},
  the agent decides valence, then arousal, then the V-A quadrant, then the best in-quadrant
  emotion---mirroring the dataset's hierarchical structure.
  \item \textbf{Post-hoc consistency.} If the chosen emotion's expected quadrant disagrees with
  the predicted $(\hat v,\hat a)$, the labels are auto-corrected to the emotion's quadrant.
  \item \textbf{Low-temperature, quality-aware decoding.} Decoding at $T{=}0.1$ with automatic
  retries on malformed output ensures valid JSON in all test cases.
\end{enumerate}
This description-first ordering is the key to grounding: by writing what it \emph{sees} before
naming what it \emph{feels}, the agent achieves a near-perfect description score of $0.9952$,
far above fine-tuned open agents whose templated rationales score $0.769$ (\S\ref{sec:exp}).

\subsection{Task Routing}
The router operationalizes the three difficulty regimes:
(i) \emph{emotion} (hard, long-tailed) is routed to the five-agent arbiter $\Pi_{\text{vote}}$;
(ii) \emph{valence/arousal} (saturated) and (iii) \emph{description} (reasoning) are routed to
the single strongest reasoning agent $A_5$, because ensembling near-saturated binary axes adds
noise rather than signal, and description quality depends on one coherent narrator rather than a
vote. This asymmetric routing---ensemble for the hard task, best-single-agent for the easy and
generative tasks---is what converts complementary agents into a high combined score.

\section{Experiments}
\label{sec:exp}

\subsection{Setup}
We train the open MLLM and vision agents on EmoArt-130k with a $90/10$ train/validation split
(stratified by emotion, later upgraded to style$\times$emotion joint stratification). Fine-tuning
uses LoRA~\cite{hu2022lora} (rank $128$, $\alpha{=}256$) via LlamaFactory~\cite{zheng2024llamafactory},
bf16, cosine schedule. We report the official validation metrics---per-task
accuracy and macro-F1 and their mean Combined score---and, for the final system, the official
held-out test-set Overall\slash Classification\slash Description scores ($1{,}000$ artworks, scored by the
challenge's multimodal LLM protocol). All agents emit structured JSON with $100\%$ parse success.
Agent-specific hyperparameters are summarized in Table~\ref{tab:hyper}.

\begin{table}[t]
\centering
\vspace{-10pt}
\caption{Agent-specific hyperparameters. Fine-tuned agents ($A_1$, $A_2$) list training settings; closed-source agents ($A_3$--$A_5$) list inference settings.}
\label{tab:hyper}
\footnotesize
\setlength{\tabcolsep}{3pt}
\begin{tabular}{@{}lp{5.8cm}@{}}
\toprule
\textbf{Agent} & \textbf{Key Hyperparameters} \\
\midrule
$A_1$ (DINOv2-G)
  & ViT-Giant/14, 2-layer MLP heads ($768$/$384$); head LR $5{\times}10^{-5}$ (backbone $0.01\times$); Mixup $\alpha{=}0.8$, CutMix $\alpha{=}1.0$; EMA $0.9998$; dropout $0.2$; label smoothing $0.1$; weight decay $0.05$; cosine + $10\%$ warmup; $15$ epochs, early stop patience $4$; batch $56$ \\
\midrule
$A_2$ (SceneCoT-30B)
  & Qwen3-VL-30B-A3B (30B total, 3B active); LR $10^{-5}$; cosine + $5\%$ warmup; $5$ epochs; batch $112$; $127{,}782$ training samples \\
\midrule
$A_3$ (GPT-5.4, prior)
  & Temp $0.3$; max tokens $800$; original image resolution \\
\midrule
$A_4$ (Gemini 3.1)
  & Temp $0.3$; max tokens $2{,}000$; image $\le\!1024$px \\
\midrule
$A_5$ (GPT-5.4, D$\to$C)
  & Temp $0.1$; max tokens $1{,}500$; image $\le\!1024$px \\
\bottomrule
\vspace{-10pt}
\end{tabular}
\end{table}

\subsection{Single-Agent Ablation: What Matters for Art Emotion?}
Table~\ref{tab:single} presents a controlled ablation study on the EmoArt-130k validation split,
grouping eleven single-agent variants into four categories:
backbone \& training strategy, sampling \& augmentation, data-side supervision, and model scaling.
Each variant highlights one key change, allowing us to isolate contributions.

\begin{table*}[t]
\centering
\vspace{-10pt}
\caption{Single-agent ablation on EmoArt-130k validation ($13.3$k artworks). Method-/scale-side changes plateau near $0.76$; decisive gains come from data-side CoT quality (SceneCoT) and scaling \emph{only with} high-quality data. Combined $=$ mean of six classification metrics. Best in \best{bold}.}
\label{tab:single}
\footnotesize
\setlength{\tabcolsep}{4.0pt}
\begin{tabular}{@{}llccccccc@{}}
\toprule
\textbf{Variant} & \textbf{Key Change} & \textbf{Emo Acc} & \textbf{Emo F1} & \textbf{Val Acc} & \textbf{Val F1} & \textbf{Aro Acc} & \textbf{Aro F1} & \textbf{Combined} \\
\midrule
\multicolumn{9}{@{}l}{\textit{\textbf{(a) Backbone \& Training Strategy}}} \\[2pt]
CoT-FT-Base      & InternVL3-8B, LoRA, frozen ViT          & 0.6972 & 0.2938 & 0.9300 & 0.8200 & 0.9033 & 0.8631 & 0.7512 \\
CoT-FT-Bal       & \quad + class balancing, diverse CoT     & 0.6849 & 0.3275 & 0.9263 & 0.8234 & 0.9047 & 0.8676 & 0.7557 \\
Qwen3-8B         & Qwen3-VL-8B backbone (baseline)          & 0.6882 & 0.3290 & 0.9311 & 0.8355 & 0.9091 & 0.8739 & 0.7611 \\
DINOv2-G         & DINOv2 ViT-Giant, vision-only            & 0.6459 & 0.3071 & 0.9225 & 0.8198 & 0.8930 & 0.8580 & 0.7411 \\
\midrule
\multicolumn{9}{@{}l}{\textit{\textbf{(b) Sampling \& Augmentation}}} \\[2pt]
FocalLoss        & Focal $\gamma{=}2$, two-stage training   & 0.6979 & 0.2791 & 0.9283 & 0.8092 & 0.9044 & 0.8657 & 0.7474 \\
AsymSample       & Aggressive asymmetric resampling         & 0.6806 & 0.3149 & 0.9320 & 0.8337 & 0.9068 & 0.8722 & 0.7567 \\
ImgAug           & Tiny-class image augmentation            & 0.7024 & 0.2953 & 0.9314 & 0.8283 & 0.9094 & 0.8727 & 0.7566 \\
VisTower         & \quad + unfreeze vision tower            & \best{0.7085} & 0.3041 & 0.9362 & 0.8382 & \best{0.9105} & 0.8736 & 0.7618 \\
\midrule
\multicolumn{9}{@{}l}{\textit{\textbf{(c) Data-Side Supervision Quality}}} \\[2pt]
SceneCoT         & Scene + real emotional-impact CoT        & 0.7005 & 0.3212 & 0.9338 & 0.8391 & 0.9085 & 0.8735 & 0.7628 \\
\midrule
\multicolumn{9}{@{}l}{\textit{\textbf{(d) Model Scaling}}} \\[2pt]
Qwen3-30B        & 30B MoE, old data                        & 0.6760 & 0.3157 & 0.9290 & 0.8328 & 0.9013 & 0.8630 & 0.7530 \\
SceneCoT-30B     & 30B MoE + SceneCoT data                  & 0.7025 & \best{0.3344} & \best{0.9375} & \best{0.8486} & 0.9099 & \best{0.8739} & \best{0.7678} \\
\bottomrule
\end{tabular}
\vspace{-10pt}
\end{table*}

\noindent\textbf{Loss and sampling interventions saturate quickly.}
Class-balanced re-sampling (CoT-FT-Bal) improves emotion macro-F1 by $+3.4$pp over the
InternVL3 baseline, but more aggressive strategies---focal loss ($\gamma{=}2$, Combined
drops to $0.7474$), asymmetric oversampling, and image-level augmentation---all fail to
beat the Qwen3-8B backbone baseline ($0.7611$). Under the extreme $1549{:}1$ class imbalance,
loss re-weighting is too blunt and high oversampling ratios ($10\times$+) cause several
tail classes to collapse to F1$=0$. Unfreezing the vision tower (VisTower) yields a modest
gain ($+0.07$pp Combined vs.\ Qwen3-8B) primarily on mid-frequency classes.

\noindent\textbf{Scale alone is insufficient.}
The $30$B MoE backbone trained on original data (Qwen3-30B, $0.7530$) \emph{underperforms} the
$8$B dense model on improved data (SceneCoT, $0.7628$), despite substantially higher training cost.
Only when the larger backbone is paired with high-quality supervision does it
reach the best single-agent score (SceneCoT-30B, $0.7678$), confirming that
\emph{data quality dominates model capacity} in this regime.

\noindent\textbf{Data-side supervision is the decisive lever.}
Replacing templated rationales with genuine scene-grounded chain-of-thought targets and
style$\times$emotion joint stratification (SceneCoT) lifts emotion macro-F1 by $+1.71$pp over the best method-side variant (VisTower) and,
critically, gives tail classes their \emph{first} non-zero F1 scores (\textsc{Aroused}
$0{\to}0.028$, \textsc{Annoyed} $0{\to}0.078$, \textsc{Tired} $0{\to}0.086$). This is the
only intervention that breaks the tail barrier without degrading head-class performance---precisely
the behavior rewarded by macro-averaged F1.

\subsection{Complementary Agents}
\begin{figure}[t]
  \centering
  \includegraphics[width=\columnwidth]{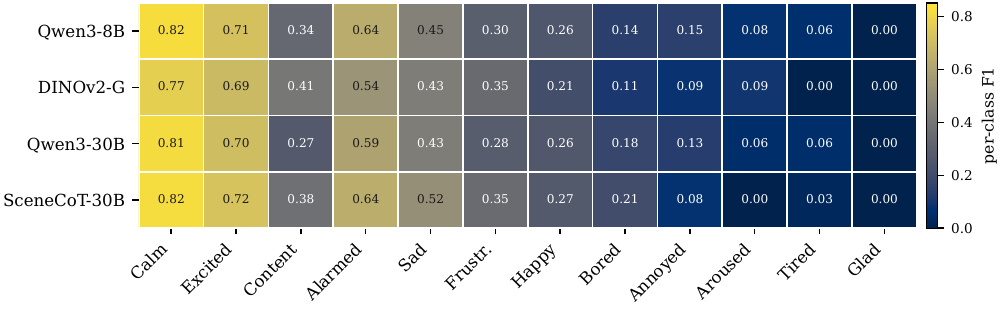}
  \caption{Per-class emotion F1 across four agents. No single agent dominates all classes, motivating the voting arbiter.}
  \label{fig:heat}
\end{figure}

Figure~\ref{fig:heat} shows per-class emotion F1 for four agents. No single agent dominates all classes: the
vision-only DINOv2 agent (DINOv2-G) is strongest on \textsc{Contentment} and \textsc{Aroused}; the
Qwen3-VL-8B agent (Qwen3-8B) leads on \textsc{Calm}, \textsc{Annoyed}, and \textsc{Tired}; while
SceneCoT-30B wins the mid-frequency classes (\textsc{Sad}, \textsc{Bored}, \textsc{Excited}).
These complementary strengths are exactly the
decorrelated errors that a vote can exploit, validating the diversity-over-capacity principle.

\subsection{Test-Set Results and the Final System}
Table~\ref{tab:test} reports both classification and description results on the
official test set. The single fine-tuned agent (SceneCoT-30B) scores well on
classification accuracy but poorly on description ($0.769$); its templated rationales lack the
visual specificity the LLM judge rewards.
Closed reasoning agents (GPT-5.4 (prior), Gemini~3.1) excel at description ($>\!0.95$) thanks
to their art-historical vocabulary and chain-of-thought reasoning, but have weaker or less
balanced classification---GPT-5.4 (prior) achieves only $0.674$ emotion accuracy because it
lacks the fine-tuned agents' distribution awareness. This complementarity is precisely what
the routing policy exploits. Our routed system---\textbf{five-agent rare-class voting for
emotion, plus the description-first reasoning agent $A_5$ for valence/arousal and all
text}---combines the best of both: the highest classification score ($0.7789$, driven by emotion
accuracy $0.796$) \emph{and} the highest description score ($0.9952$), for an
\textbf{Overall Score of $0.8870$}, well above any single agent. The $0.226$-point gap between
the fine-tuned model's description ($0.769$) and $A_5$'s ($0.995$) confirms that grounded
description is a reasoning capability that benefits more from prompt engineering than from
supervised fine-tuning on this dataset.

\begin{table*}[t]
 \vspace{-6pt}
\centering
\caption{Official Track-2 test results ($1{,}000$ artworks). Emotion is a 12-class task; V/A are binary. Description scored $[0,1]$ by the challenge LLM judge. Emotion routed to the 5-way voter; V/A and description to $A_5$.
Best in \best{bold}.}
\label{tab:test}
\footnotesize
\setlength{\tabcolsep}{4.0pt}
\begin{tabular}{@{}l cc cc cc c c ccc c c@{}}
\toprule
\multirow{2}{*}{System}
  & \multicolumn{2}{c}{Emotion} & \multicolumn{2}{c}{Valence} & \multicolumn{2}{c}{Arousal}
  & \multirow{2}{*}{\shortstack{Avg\\F1}} & \multirow{2}{*}{\shortstack{Cls.\\Score}}
  & \multirow{2}{*}{\shortstack{Vis.\\Ground.}} & \multirow{2}{*}{\shortstack{Attr.\\Spec.}} & \multirow{2}{*}{\shortstack{Overall\\Cap.}}
  & \multirow{2}{*}{\shortstack{Desc.\\Score}}
  & \multirow{2}{*}{\shortstack{\textbf{Overall}\\\textbf{Score}}} \\
\cmidrule(lr){2-3}\cmidrule(lr){4-5}\cmidrule(lr){6-7}
& Acc & F1 & Acc & F1 & Acc & F1 & & & & & & & \\
\midrule
$A_2$: SceneCoT-30B (FT)  & 0.768 & 0.221 & 0.868 & 0.729 & 0.935 & 0.865 & 0.605 & 0.731 & 0.822 & 0.656 & 0.830 & 0.769 & 0.750 \\
$A_3$: GPT-5.4 (prior)   & 0.674 & 0.269 & 0.897 & 0.826 & 0.930 & 0.860 & 0.652 & 0.743 & 0.967 & 0.963 & 0.976 & 0.969 & 0.856 \\
$A_4$: Gemini 3.1        & 0.763 & \best{0.309} & 0.883 & 0.813 & 0.917 & 0.851 & 0.658 & 0.756 & 0.955 & 0.950 & 0.959 & 0.955 & 0.855 \\
$A_5$: GPT-5.4 (D$\to$C) & 0.614 & 0.273 & \best{0.903} & \best{0.858} & \best{0.937} & \best{0.874} & 0.668 & 0.743 & \best{0.995} & \best{0.996} & \best{0.995} & \best{0.995} & 0.869 \\
3-way majority vote    & 0.741 & 0.301 & -- & -- & -- & -- & -- & -- & -- & -- & -- & -- & -- \\
\textbf{\sys{} (ours)} & \best{0.796} & 0.305 & \best{0.903} & \best{0.858} & \best{0.937} & \best{0.874} & \best{0.679} & \best{0.779} & \best{0.995} & \best{0.996} & \best{0.995} & \best{0.995} & \best{0.887} \\
\bottomrule
\end{tabular}
\vspace{-3pt}
\end{table*}

\subsection{Ablations}
\noindent\textbf{Voting arbiter.} Table~\ref{tab:test} (row 5 vs.\ the final row) isolates the emotion controller on the
test set (description and V/A held fixed). The \emph{3-way vote} uses three closed-source agents
($A_3$, $A_4$, $A_5$) with simple majority.
The \emph{5-way vote} (our final system) extends this to all five agents---$A_3$, $A_5$,
$A_4$, $A_2$, and $A_1$---and introduces
a \emph{rare-class priority} rule ($\tau_r{=}2$): whenever $\ge\!2$ voters agree on a tail-class
label (\texttt{happy}, \texttt{aroused}, \texttt{glad}, \texttt{annoyed}, \texttt{tired}), that
label is adopted even if the majority predicts a head class. This mechanism lifts accuracy by
$+5.5$pp over plain 3-way voting, confirming that it is the rare-class boost, not voting alone,
that captures the tail.

\noindent\textbf{Description-first prompting.} Comparing $A_3$ (GPT-5.4, label-first prior)
with $A_5$ (GPT-5.4, D$\to$C) in Table~\ref{tab:test} isolates the effect of prompt ordering
on the \emph{same} backbone: description score rises from $0.969$ to $0.995$ (+2.6pp), with
attribute specificity jumping from $0.963$ to $0.996$. The gap widens further against the
fine-tuned $A_2$ (description $0.769$, specificity $0.656$), confirming that writing visual
evidence \emph{before} the label is the key driver of grounded, specific descriptions.

\noindent\textbf{Routing.} Replacing the routed valence/arousal (single reasoning agent) with a
five-agent vote leaves V/A essentially unchanged ($\le\!0.2$pp) while adding latency---consistent
with the saturation analysis of \S\ref{sec:problem}: ensembling near-saturated binary axes is not
worthwhile, so routing them to a single agent is both simpler and as accurate.

\begin{figure*}[t]
\centering
\setlength{\abovecaptionskip}{4pt}
\setlength{\belowcaptionskip}{0pt}
\begin{minipage}[t]{0.33\textwidth}\centering
\caseimg{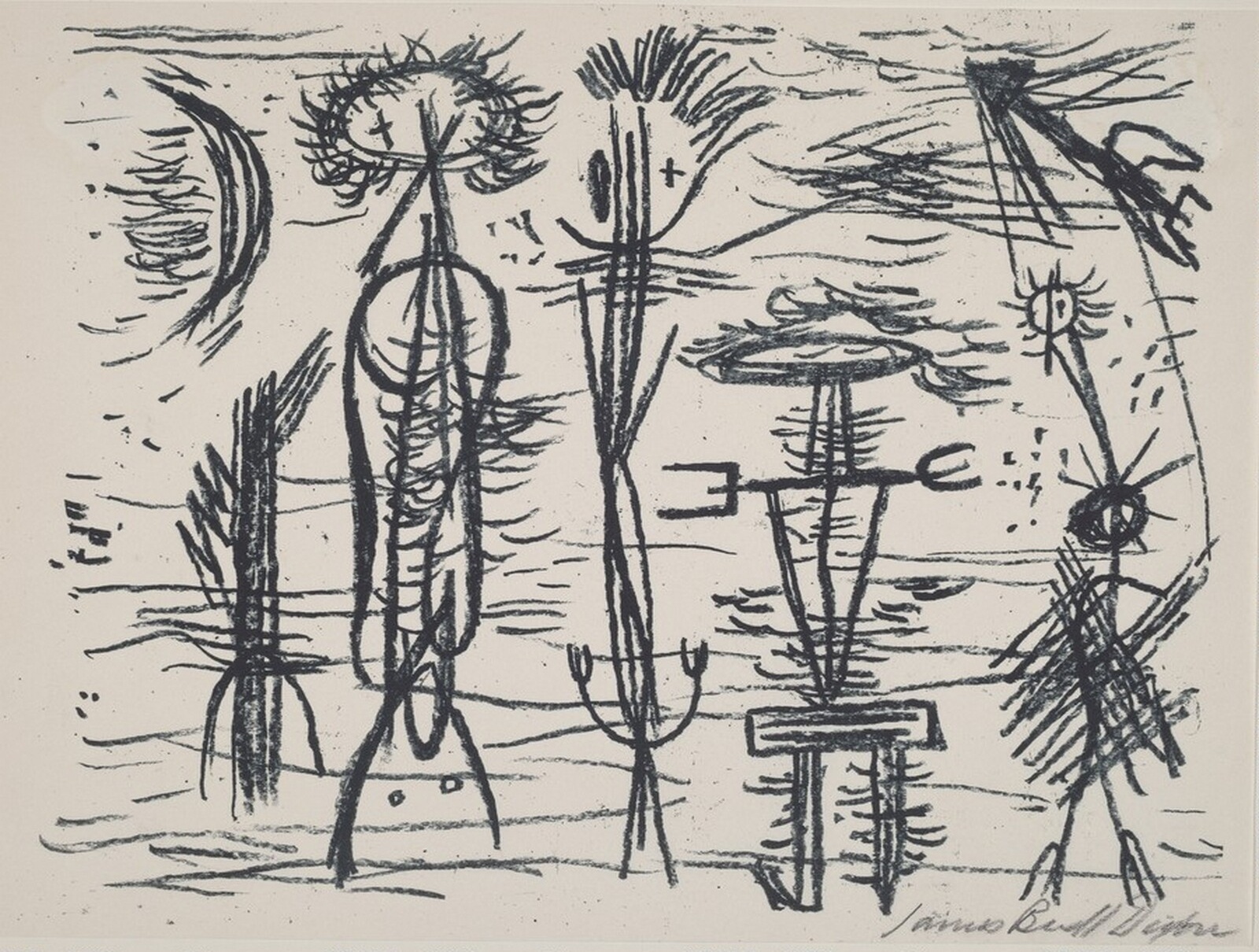}\\[1pt]
{\scriptsize\sffamily\bfseries\textcolor[HTML]{e65100}{Alarmed} \textbar\ Negative / High}\\[1pt]
\casetxt{%
\textbf{Overall:} A dramatic scene charged with confrontation and physical conflict, creating a strong sense of danger and urgency.
\textbf{Brushstroke:} Bold, rapid strokes convey explosive force and kinetic energy, amplifying the sense of violence.
\textbf{Color:} Intense contrasts with dark tones and sharp highlights heighten visual tension and emotional alarm.
\textbf{Composition:} Figures in dynamic, clashing poses dominate the frame with strong diagonals, creating instability and chaos.
\textbf{Line:} Sharp, angular, and agitated lines define forceful gestures and generate nervous, confrontational energy.
\textbf{Light:} High-contrast illumination with stark tonal shifts intensifies the drama and sense of crisis.}
\end{minipage}\hfill
\begin{minipage}[t]{0.33\textwidth}\centering
\caseimg{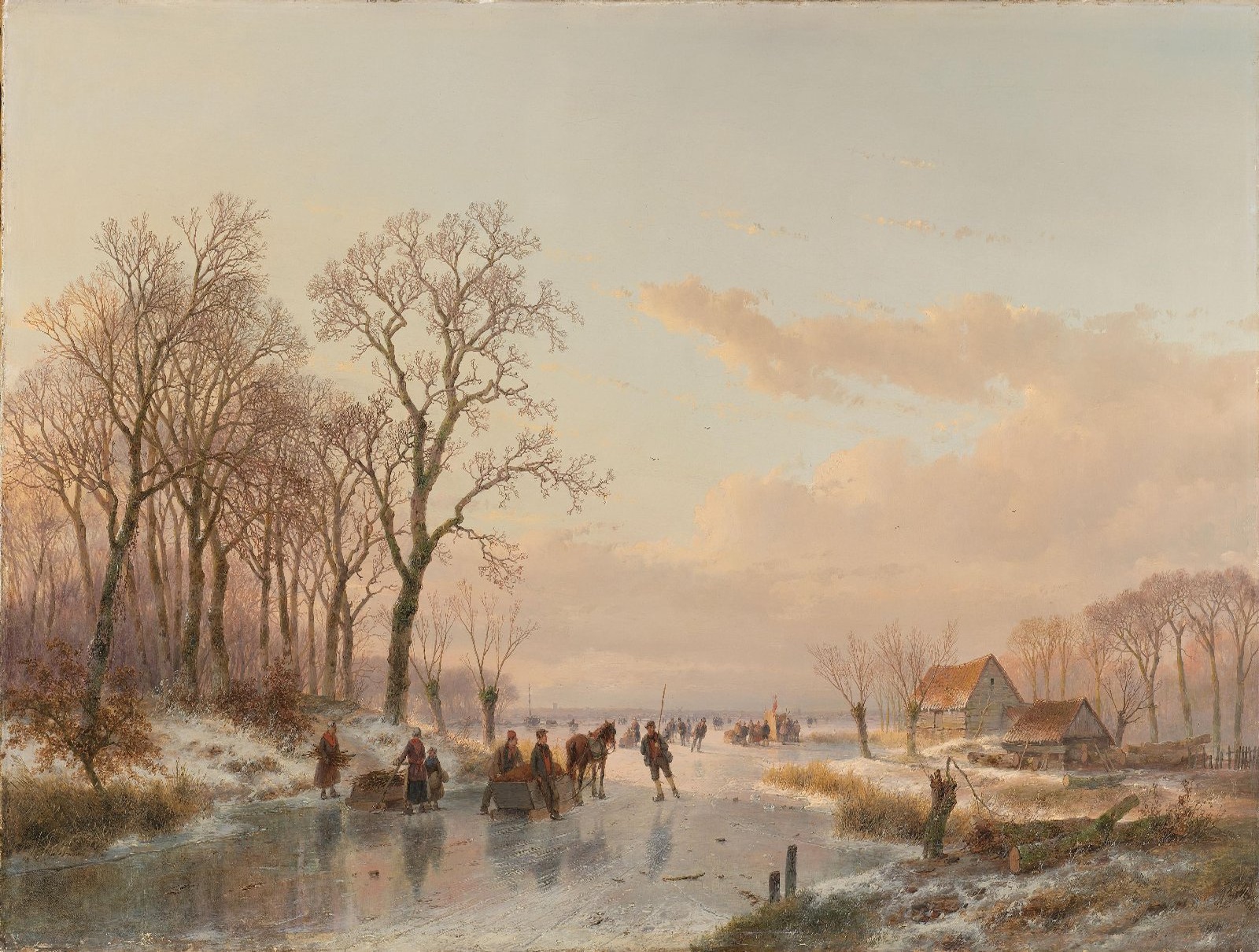}\\[1pt]
{\scriptsize\sffamily\bfseries\textcolor[HTML]{c62828}{Excited} \textbar\ Positive / High}\\[1pt]
\casetxt{%
\textbf{Overall:} A vibrant scene filled with dynamic activity, social interaction, and festive energy, creating a mood of cheerful excitement.
\textbf{Brushstroke:} Lively and varied, with energetic marks that support the bustling visual activity.
\textbf{Color:} Warm and varied palette with reds, golds, and bright accents that energize the scene and evoke celebration.
\textbf{Composition:} Densely populated with figures and activity distributed across the frame, creating constant visual incident.
\textbf{Line:} Active, rhythmic, and decorative, defining figures and objects with clarity and animation.
\textbf{Light:} Bright and even, with warm illumination that makes the scene feel open, inviting, and energized.}
\end{minipage}\hfill
\begin{minipage}[t]{0.33\textwidth}\centering
\caseimg{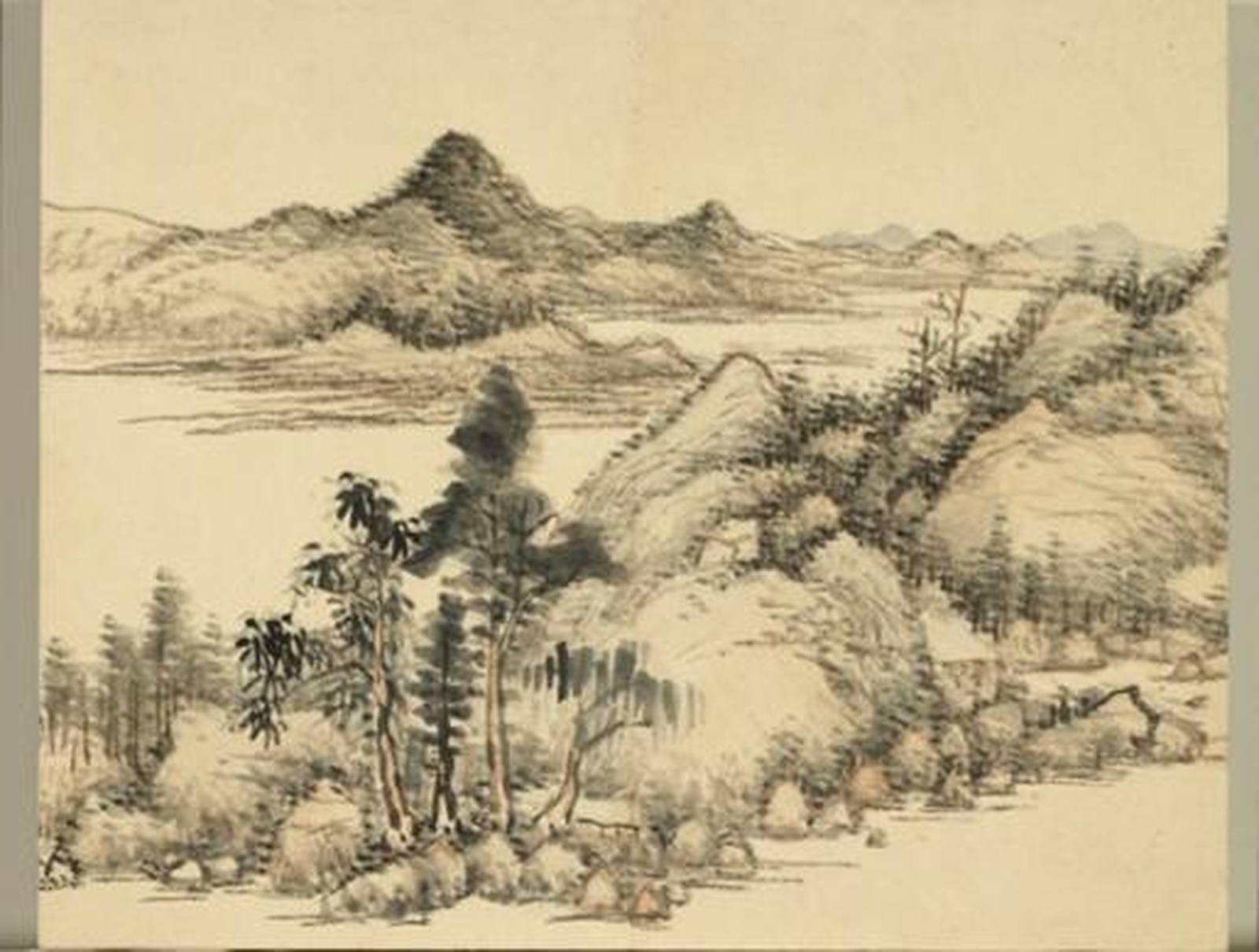}\\[1pt]
{\scriptsize\sffamily\bfseries\textcolor[HTML]{2e7d4f}{Calm} \textbar\ Positive / Low}\\[1pt]
\casetxt{%
\textbf{Overall:} A quiet ink landscape of mountains and water rendered with restraint and spaciousness, creating serene contemplation.
\textbf{Brushstroke:} Delicate and controlled, with light dry-brush textures describing hills and trees; meditative and settled.
\textbf{Color:} Extremely subdued---black ink washes on warm beige paper with only faint tonal variation; gentle and understated.
\textbf{Composition:} Balanced across foreground trees, middle-ground slopes, and distant mountains with generous negative space.
\textbf{Line:} Fine, organic, and slightly wavering, shaping ridges and foliage without sharp angular force; calm rhythm.
\textbf{Light:} Diffuse and even, with no harsh contrasts; soft illumination contributes to a tranquil, breathable atmosphere.}
\end{minipage}
\caption{Qualitative results on three test artworks showing the emotion, V/A quadrant, and attribute-grounded description.}
\label{fig:cases}
\end{figure*}

\subsection{Qualitative Results}
\label{sec:qual}
Figure~\ref{fig:cases} shows representative test artworks spanning diverse emotions and
V/A quadrants. The descriptions are grounded in concrete visual evidence---palette,
composition, lighting---rather than generic affect words. This evidence-before-label
behavior is exactly what the \dtoc{} ordering elicits, explaining the near-perfect
description score in Table~\ref{tab:test}.

\subsection{Post-Processing Pipeline}
Every agent response passes through a deterministic
pipeline; fewer than $2\%$ of test responses require correction.
\textbf{(i)~JSON repair:} markdown fencing is stripped and, on parse failure, the
outermost JSON span is extracted; missing fields trigger one re-query.
\textbf{(ii)~Label normalization:} variants are mapped to canonical emotions
(e.g., ``contentment''$\to$``content''), and unresolved labels are inferred from the
predicted V/A quadrant.
\textbf{(iii)~V/A consistency:} we enforce the deterministic emotion-to-V/A mapping of
Russell's circumplex~\cite{russell1980circumplex}, replacing conflicting binary outputs
with the mapping-consistent pair.
\textbf{(iv)~Text fallback:} empty fields are filled with a short generic template to guarantee a valid submission format.

\subsection{System Prompt for Agent $A_5$}
\label{app:prompt}

The description-first reasoning agent $A_5$ (GPT-5.4, D$\to$C) uses a structured system
prompt with three blocks: (1)~role definition, (2)~emotion taxonomy with pairwise
differentiation, and (3)~JSON output schema enforcing DESCRIBE$\to$CLASSIFY order.
Each emotion lists its visual cues and contrasts with confusable neighbors (e.g.,
\textsc{Calm} \emph{vs.}\ \textsc{Bored}: ``calmness has beauty and intention''), and
\texttt{emotion\_reasoning} must name the quadrant and argue among three candidates.
Crucially, the cues are organized along the same axes the task asks us to describe---palette,
line, composition, lighting---and the prompt requires each attribute to be tied to its
emotional effect rather than merely listed, giving the descriptions a shared evidence
vocabulary with the classification rationale.

\noindent\textbf{Per-image user prompt.}
Each image is accompanied by: ``\textit{Analyze this artwork's emotional dimensions.
Step~1: Observe brushwork, composition, colors, lines, lighting and the atmosphere.
Step~2: Classify---determine valence (Positive/Negative), arousal (High/Low), reason through
the quadrant candidates, and give your final emotion. Output as JSON with descriptions FIRST,
then classifications.}''

\section{Conclusion}

We addressed the AffectiveArt 2026 Track-2 challenge by recognizing that its four sub-tasks
live in incompatible difficulty regimes---an extreme long-tail, near-saturated binary axes,
and a generative reasoning task. Instead of one monolithic model, we built \sys{}, an
\textbf{agentic society of heterogeneous experts} coordinated by a rare-class-aware voting
arbiter and a description-first reasoning agent, with a routing policy matching each sub-task
to the right agent configuration, reaching an \textbf{Overall Score of $0.8870$} on the
official test set. An eleven-variant ablation showed that, past method- and scale-side
saturation, the decisive gains come from \emph{agent collaboration} and \emph{data-side
supervision} rather than larger backbones---a $30$B MoE model trained on older data does not
beat an $8$B model on better data. The transferable lesson for multimedia affective tasks:
when a benchmark bundles heterogeneous objectives, designing the \emph{society of agents and
their routing} can matter more than scaling any single one.



\begin{acks}
This work was supported by the Special Fund for Key Program of Science and Technology of Jiangsu Province (No.\ BG2024042) and the Gusu Leading Talents Program for Innovation and Entrepreneurship (No.\ ZXL2025323).
\end{acks}

\bibliographystyle{ACM-Reference-Format}
\bibliography{sigconf-authordraft.bib}

\end{document}